\PassOptionsToPackage{table,xcdraw}{xcolor}
\documentclass[runningheads]{llncs}

\usepackage{eccvabbrv}
\usepackage{graphicx}
\usepackage{booktabs}
\usepackage[accsupp]{axessibility}
\usepackage{amsmath,amssymb}
\usepackage{mathtools}

\usepackage{amsthm}
\usepackage{multirow}
\usepackage{enumitem}
\usepackage{algorithm}
\usepackage{algpseudocode}
\usepackage{xspace}
\usepackage{tikz}
\usepackage{pgfplots}
\usepackage{authblk}

\pgfplotsset{compat=1.18}
\definecolor{acccolor}{RGB}{33,113,181}
\definecolor{latcolor}{RGB}{217,95,2}

\definecolor{acccolor}{RGB}{33,113,181}   
\definecolor{latcolor}{RGB}{217,95,2}     

\usepackage{hyperref}

\hypersetup{
    colorlinks=true,      
    linkcolor=red,         
    citecolor=blue,        
    urlcolor=purple         
}

\usepackage{hyperref}

\usepackage{orcidlink}

\usepackage{tcolorbox}
\usepackage{float}
\usepackage{tikz} 
\usepackage{xspace}
\usepackage{etoc}
\usepackage{pifont}

\newcommand{\ourmethod}{SeamCam\xspace}
\newcommand{\ourdataset}{CamFG-1.5K\xspace}

\theoremstyle{plain}

\theoremstyle{remark}

\begin{document}

\title{
\ourmethod: Quantifying Seamless Camouflage via Multi-Cue Visual Detectability}

\titlerunning{SeamCam}





\author{Amin Karimi Monsefi\inst{1}\textsuperscript{\dag} \and
Abolfazl Meyarian\inst{2}\textsuperscript{\dag} \and
Mridul Khurana\inst{3} \and 
Shuheng~Wang\inst{1} \and
Pouyan Navard\inst{1}\textsuperscript{\ddag} \and
Cheng Zhang\inst{4} \and
Anuj Karpatne\inst{3} \and 
Wei-Lun~Chao\inst{5} \and 
Rajiv Ramnath\inst{1}}


\authorrunning{A.K.~Monsefi et al.}


\institute{
\(^{\text{1}}\)The Ohio State University \hspace{0.2cm} 
\(^{\text{2}}\)Path Robotics, USA \hspace{0.2cm} 
\(^{\text{3}}\)Virginia Tech \hspace{0.2cm} \\
\(^{\text{4}}\)Texas A\&M University \hspace{0.2cm} 
\(^{\text{5}}\)Boston University
}

\maketitle
\let\thefootnote\relax
\footnotetext{\textsuperscript{\dag} Equal contribution.}
\footnotetext{\textsuperscript{\ddag} PhD Alumni, The Ohio State University; currently working in industry.}

\begin{abstract}
Animals are described as effectively camouflaged when they blend seamlessly with their surrounding, yet no standardized quantitative measure of this seamlessness exists. We address this gap by framing camouflage evaluation as a visual localization problem: a well-camouflaged animal is one that remains difficult to detect even when its category is known. We introduce \textbf{SeamCam}\footnote{Project page: \url{https://7amin.github.io/SeamCam/}} \textbf{(Seam}less \textbf{Cam}ouflage), a metric that quantifies how detectable an animal is from the available visual evidence. Given an image and a target species, SeamCam generates category-conditioned detection proposals, extracts segmentation masks, and identifies the subset whose collective union yields the highest IoU with the ground-truth mask — the SeamCam score is one minus this maximum recoverable localization signal, where a higher score indicates stronger camouflage (i.e., lower detectability). In a human two-alternative forced-choice study with 94 participants and 2,390 comparisons, SeamCam achieves \textbf{78.82\%} agreement with human camouflage difficulty judgments, substantially outperforming state-of-the-art by a margin of \textbf{$\sim25\%$}. We then demonstrate \ourmethod's utility as a preference signal for Direct Preference Optimization (DPO) to fine-tune a diffusion-based inpainting model for camouflage generation. This offers an affordable training approach with an objective explicitly suited for camouflage generation, unlike that of typical diffusion models. Additionally, to support rigorous benchmarking of camouflage generation, we further introduce \textbf{CamFG-1.5k}, a curated dataset of \textbf{1,521} high-resolution images in which animals are fully visible prior to camouflage generation, enabling unbiased evaluation by controlling for occlusion artifacts present in existing datasets.
\end{abstract}    
\section{Introduction}

Animal camouflage is a visually grounded phenomenon that has long been described in intuitive terms, often characterized by an animal’s ability to blend seamlessly into its surrounding environment. In natural settings, observers frequently describe camouflaged animals as being “hard to see,” “visually integrated,” or “indistinguishable from the background,” suggesting that effective camouflage arises when the animal does not perceptually stand out as a separate object. This notion of seamless blending reflects a perceptual experience rather than a specific measurable property, yet it plays a central role in how camouflage is discussed across biology, ecology, and computer vision. The prevalence of such language highlights the importance of understanding camouflage as a holistic visual phenomenon, motivating the need for a principled way to formalize and study visual seamlessness.

\begin{figure}[t]
    \centering
    \includegraphics[width=1\textwidth]{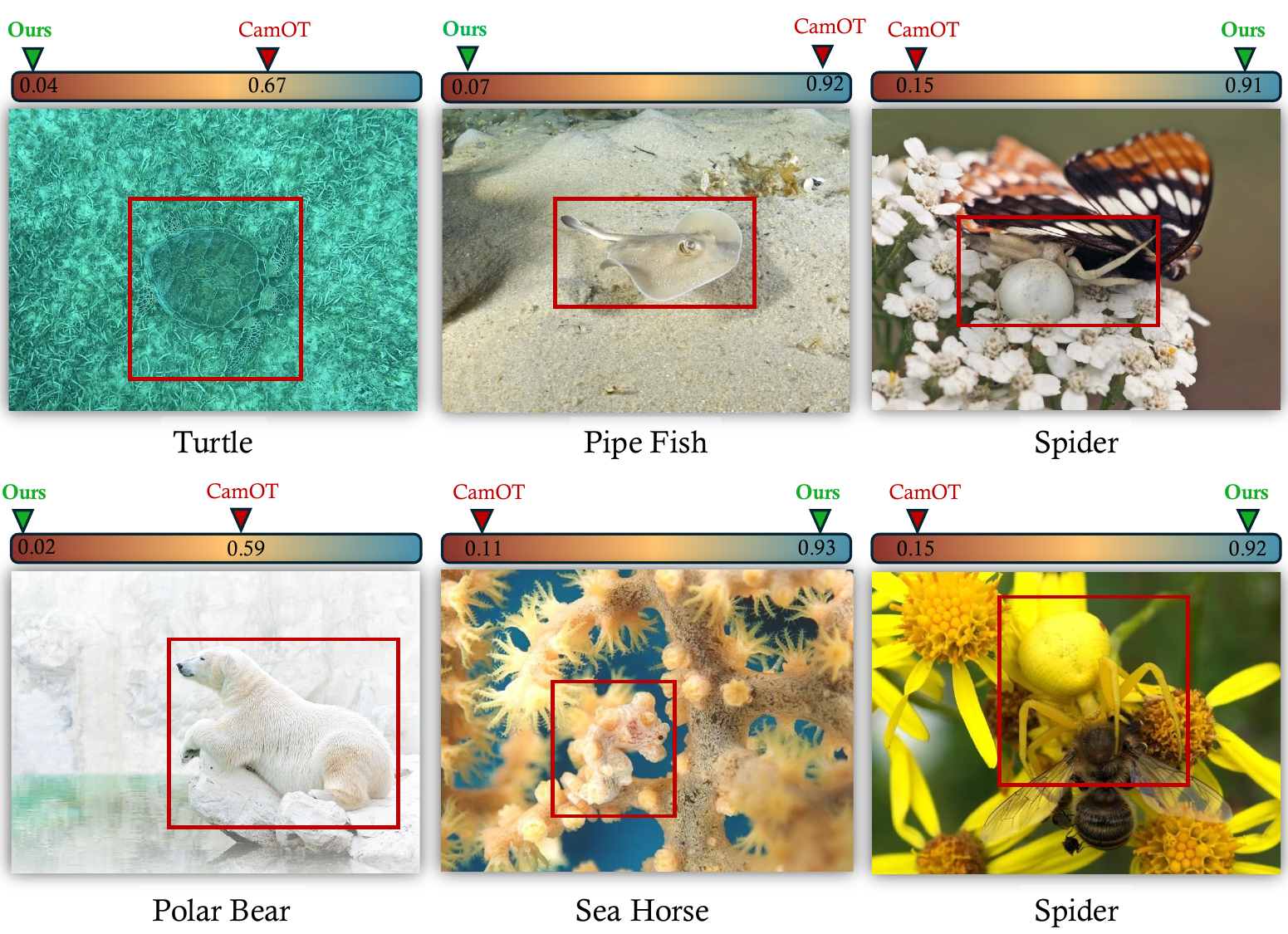}
    \vspace{-4mm}
    \caption{\textbf{\ourmethod \emph{vs.}~CamOT}. SeamCam produces consistent and accurate camouflage difficulty scores across diverse scenarios, whereas CamOT exhibits notable inconsistencies. As illustrated by the polar bear — where superficial color and lighting similarity between subject and background causes CamOT to erroneously overestimate camouflage effectiveness despite the subject being plainly visible — CamOT assigns a disproportionately high score relative to SeamCam (where higher scores indicate harder-to-detect camouflage in both metrics), revealing a fundamental limitation of its background-foreground distance-based formulation.}
    \label{fig:metrics_diff}
    \vspace{-3mm}
\end{figure}

\begin{figure}[t]
    \centering
    \includegraphics[width=1\textwidth]{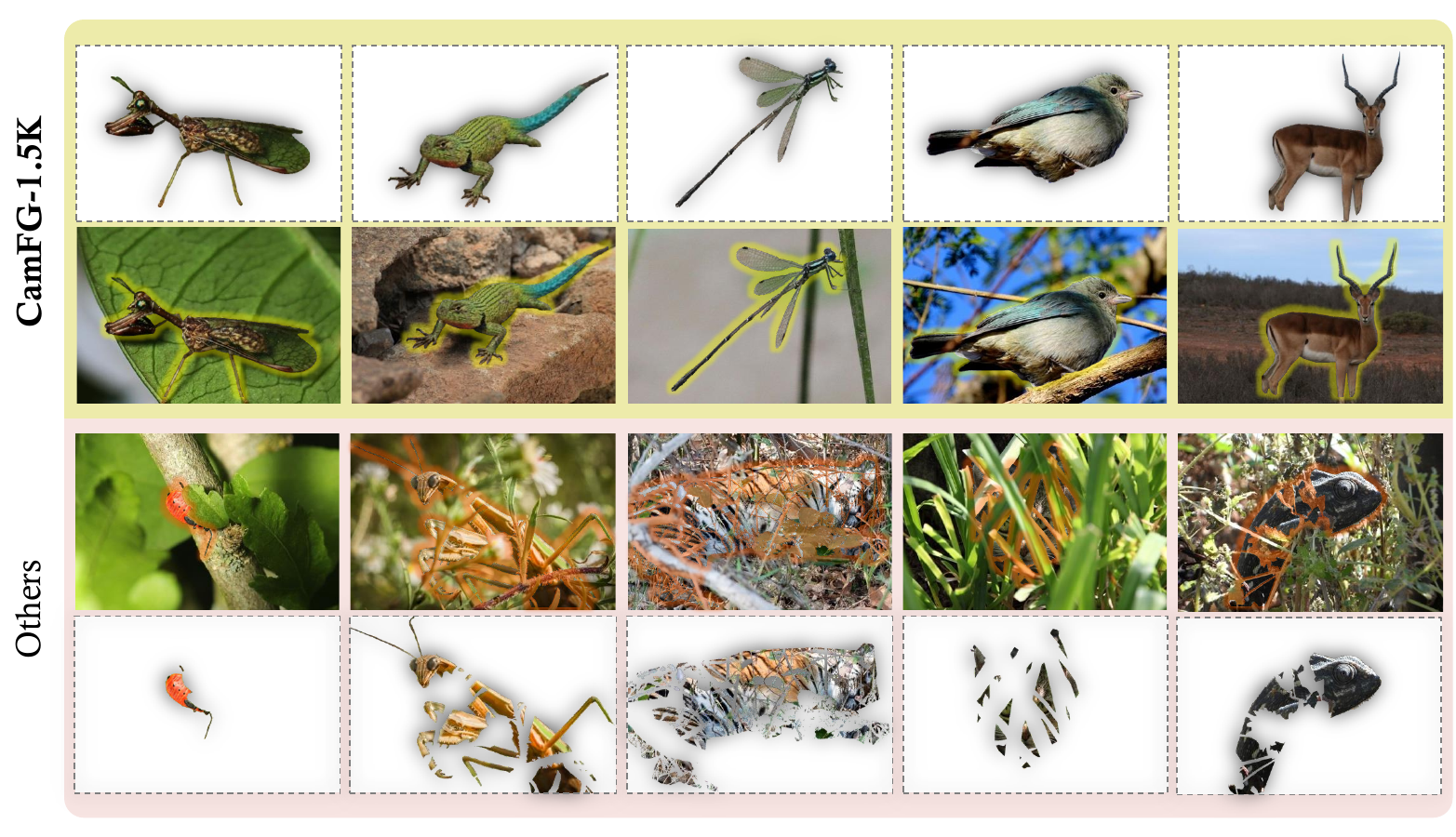}
    \vspace{-5mm}
    \caption{\textbf{Quality comparison between \ourdataset \emph{vs.} existing datasets}. Existing datasets often include subjects that are cropped, or partially camouflaged, leading to biased evaluation of camouflage models. In contrast, \ourdataset features clearly visible animals with minimal obstructions, enabling unbiased model assessment.}
    \label{fig:camfg_dataset}
    \vspace{-3mm}
\end{figure}

Camouflage is fundamentally a problem of perception rather than appearance alone. Biological studies suggest that visual recognition does not rely on a single cue but instead emerges from an iterative process in which observers evaluate multiple competing hypotheses about the content of a scene under both sensory evidence and prior expectations~\cite{merilaita2017camouflage}. From this perspective, successful camouflage does not simply minimize low-level differences between a foreground object and its background; rather, it disrupts the accumulation of visual evidence that would allow an observer to confidently infer that a target is present — operating by sustaining ambiguity across multiple plausible interpretations of a scene rather than by maximizing similarity under a fixed criterion. However, most existing computational camouflage scoring methods conflict with this view by implicitly reducing effectiveness to a single static similarity measure. For example, CamOT~\cite{das2025camouflage} reduces camouflage to a global distance between foreground and background feature distributions via optimal transport~\cite{montesuma2024recent}, while Context-Measure~\cite{wang2024context} relies on probabilistic pixel-wise correlations. By collapsing evaluation to a single comparison, these methods overlook the fact that camouflage is not a binary property, nor reducible to a single similarity measure, but instead reflects \textit{how easily competing localization hypotheses can be resolved}. Motivated by this view, we frame camouflage detectability as the difficulty of visually localizing an animal within a scene: effective camouflage prolongs or suppresses this process by fragmenting evidence across multiple weak cues rather than eliminating any single cue entirely, naturally lending itself to an image-based formulation in which camouflage efficacy is quantified by the best achievable localization performance under all plausible interpretations of the scene.

Based on this formulation, we introduce \textbf{\ourmethod}, a quantitative camouflage score derived from a visual localization framework. Given an image and an animal category, we generate a set of category-conditioned candidate detections and their corresponding segmentation masks, representing multiple plausible hypotheses about the animal’s location. Rather than selecting a single detection or mask, \ourmethod evaluates \textit{all} subsets of these proposals, measuring how well each subset collectively explains the true animal extent. In scenes where an animal is poorly camouflaged, a small number of strong cues suffice to localize it accurately, leading to high detectability and a low \ourmethod score. In contrast, when camouflage is effective, evidence is distributed across many weak and ambiguous cues, and even the best aggregation of hypotheses fails to recover the animal’s full extent. By scoring camouflage according to the strongest achievable localization rather than average similarity or pixel-level agreement, \ourmethod measures how camouflage seamlessly blends with the scene as shown in Figure~\ref{fig:metrics_diff}.


In summary, this paper makes \textbf{three} primary contributions. First, we introduce \textbf{\ourmethod}, a quantitative camouflage metric that formulates camouflage efficacy in terms of proposal-based visual localization, explicitly accounting for the integration of multiple candidate cues rather than relying on single global or pixel-level similarity measures. Unlike existing metrics, \ourmethod is grounded in how humans perceive camouflage: in a 2AFC study with 94 participants and 2,390 pairwise comparisons, \ourmethod achieves \textbf{78.82\%} agreement with human judgments, outperforming the strongest baseline by \textbf{$\sim$25\%} and demonstrating substantially stronger perceptual alignment. Second, we demonstrate that \ourmethod serves as an effective preference signal for Direct Preference Optimization, enabling diffusion-based inpainting models to synthesize more perceptually convincing camouflage than standard training objectives or alternative scoring criteria. Concretely, DPO fine-tuning with \ourmethod-based preference selection outperforms differnt variants of inpainting vbaselines across all metrics. Third, we present \textbf{\ourdataset}, a curated dataset of 1,521 high-quality samples in which animals are fully visible prior to camouflage generation. Existing camouflage datasets contain images that are already partially camouflaged, cropped, or occluded — confounds that systematically bias generation evaluation; \ourdataset eliminates these confounds by construction, enabling rigorous and unbiased benchmarking of camouflage generation models, as illustrated in Figure~\ref{fig:camfg_dataset}.
\section{Related Work}
\label{sec:related_work}

\subsection{Camouflage Evaluation Metrics}

\begin{figure*}[t]
    \centering
    \includegraphics[width=\textwidth]{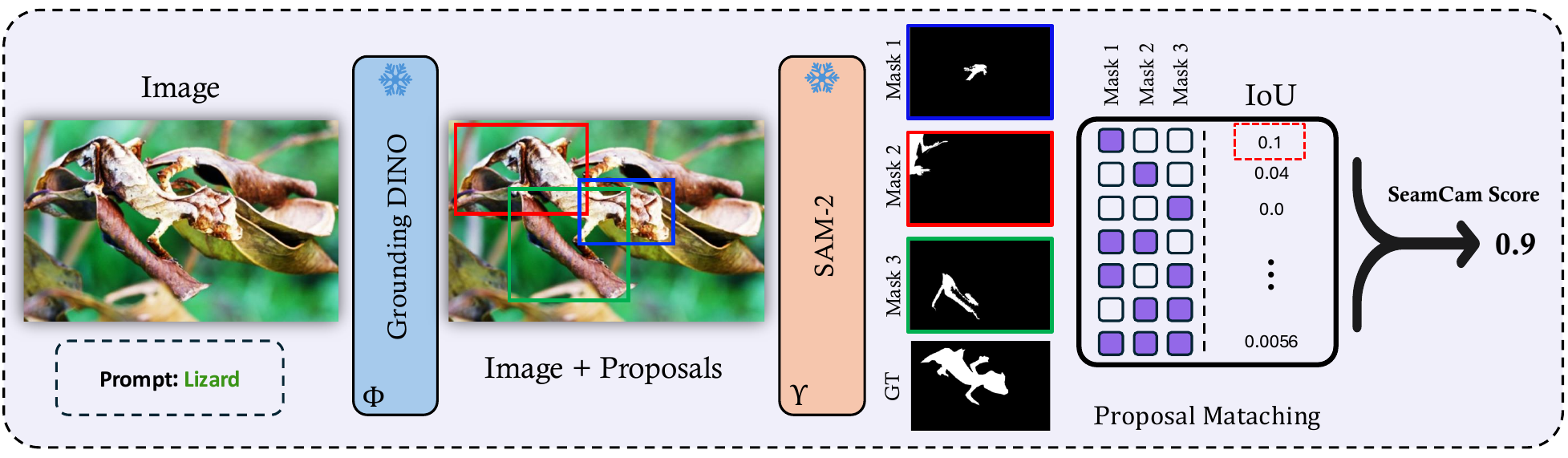}
    \vspace{-6mm}
    \caption{\textbf{Overview of \ourmethod framework.} Given an image and species name, we generate category-conditioned detection proposals via GroundingDINO, apply semantic and confidence gating, and obtain segmentation masks from SAM-2. We then evaluate \emph{all} proposal subsets, computing IoU between each subset's mask union and the ground truth. The maximum achievable IoU defines detectability $D$; the camouflage score is $1-D$. This formulation captures multi-cue integration for camouflage efficacy score measurement.}
    \label{fig:method_overview}
    \vspace{-3mm}
\end{figure*}

Standard image generation quality metrics such as FID~\cite{heusel2017gans} and KID~\cite{binkowski2018demystifying} are inherently agnostic to the concept of camouflage, as they assess global distributional similarity rather than the perceptual difficulty of detecting a foreground object within a scene. As a result, only a small number of works have proposed metrics that explicitly aim to quantify the camouflage effectiveness of a foreground object in an image. Among these, CamOT~\cite{das2025camouflage} measures camouflage by computing an optimal transport distance between foreground and background feature distributions, assigning higher scores to images with greater statistical similarity. While intuitive, this formulation conflates texture or feature matching with detectability, and may assign high camouflage scores to visually cluttered scenes in which the target remains easy to identify once its category is known. More recently, Context-measure~\cite{wang2024context} incorporates spatial context into camouflage evaluation by extending existing segmentation metrics to account for foreground–background relationships. Such metrics remain fundamentally grounded in pixel-level correspondence which limits their ability to capture the sustained perceptual ambiguity that characterizes effective camouflage. In contrast to these approaches, we introduce \textbf{\ourmethod}, a camouflage evaluation metric that quantifies camouflage through the difficulty of visually localizing an object by aggregating multiple distributed cues within the scene.

\subsection{Camouflage Image Generation}

Early work on camouflage image generation primarily relied on classical image editing techniques such as alpha blending and Poisson editing~\cite{perez2023poisson,yu2019nighttime,barrett2002object,farbman2009coordinates,hussain2015efficient,morel2012fourier,leventhal2006poisson,bie2011free}, texture transfer~\cite{efros2023image,ashikhmin2003fast,wang2022texture,lee2010directional,elad2017style}, or combinations thereof. These approaches attempt to conceal a foreground object by transferring low-level appearance statistics from the surrounding background while preserving coarse object structure. For instance,~\cite{chu2010camouflage} formulates camouflage as the task of hiding a foreground figure while maintaining partial recognizability through luminance assignment and texture synthesis. Such methods are highly sensitive to appearance discrepancies, boundary conditions, and often require substantial manual intervention to achieve optimal results. Motivated by these limitations, subsequent learning-based approaches were introduced to improve adaptability and realism. For example,~\cite{zhang2020deep} proposes a neural style transfer framework inspired by visual perception mechanisms, while LCG-Net~\cite{li2022location} employs a VGG-19–based encoder–decoder architecture to adaptively fuse foreground and background features using point-to-point structural similarity. In parallel, adversarial camouflage methods such as AdvCam~\cite{duan2020adversarial} aim to generate naturally appearing images that intentionally induce failures in recognition systems.

More recently, diffusion-based generative models~\cite{huang2023region,andang10modified,zhang2020deep,hatamizadeh2024diffit,monsefi2025taxadiffusion,croitoru2023diffusion,nie2026transition,daras2024warped,tian2025audiox,navard2024knobgen,khurana2026taxaadapter} have emerged as a powerful paradigm for camouflage image generation, offering substantial improvements in realism and flexibility. Latent Diffusion Models (LDM)~\cite{rombach2022high,li2023gligen,khurana2024hierarchical} perform diffusion in the latent space of pretrained autoencoders and incorporate cross-attention with far lower computational cost than pixel-space diffusion models such as~\cite{ho2020denoising,songdenoising}. T-Fill~\cite{zheng2022bridging} formulates image completion as a directionless sequence-to-sequence task and uses a transformer with non-overlapping CNN receptive fields and an attention-aware layer to better capture long-range context for large-mask inpainting. RePaint~\cite{lugmayr2022repaint} performs free-form image inpainting by using a pretrained unconditional Denoising Diffusion Probabilistic Model and conditioning the reverse diffusion process with known image regions, enabling high-quality and diverse results across arbitrary mask types without retraining the model. LAKE-RED~\cite{zhao2024lake} introduces a retrieval-augmented diffusion framework for background-free camouflage generation, while Camouflage Anything~\cite{das2025camouflage} proposes controlled out-painting through representation engineering to achieve seamless concealment. In this paper, we demonstrate through extensive experiments that \textbf{\ourmethod} provides an effective preference signal for training camouflage generation models via Direct Preference Optimization, enabling inpainting models to produce more realistic and perceptually convincing camouflage than those trained with standard diffusion objectives.

\section{Method}
\label{sec:method}

\subsection{\ourmethod}

Humans do not detect camouflaged animals through a single decisive cue. Instead, visual search typically accumulates multiple weak pieces of evidence distributed across the scene—partial contours, subtle texture inconsistencies, fragmented shape boundaries, or slight semantic anomalies. These cues may be individually insufficient for confident detection, yet collectively allow the observer to localize the target. The perceived difficulty of camouflage reflects how easily such distributed evidence can be integrated into a coherent object hypothesis.

Existing computational metrics often reduce this sophisticated perceptual process to a single comparison between foreground and background statistics. For example, CamOT~\cite{das2025camouflage} models camouflage via optimal transport distance between foreground and background feature distributions, thereby introducing a proxy based on global similarity. While mathematically well-defined, such formulations collapse multi-cue inference into a single scalar distance and do not explicitly model the integration of fragmented object-level evidence. In contrast, our goal is to quantify camouflage in a manner that mirrors the accumulation and aggregation of distributed visual cues.

To operationalize this idea, we formalize camouflage estimation as a category-aware localization difficulty function. Let $I \in \mathbb{R}^{H \times W \times 3}$ denote an image containing an instance of a species $c$, with corresponding ground-truth masks $M = \{ M_j \}_{j=1}^{m}$. We denote \textit{\ourmethod} by $\zeta$ and we seek the following function: $\zeta(I, M, c) \in [0,1]$, that quantifies the difficulty of localizing the target species within the image. The desired score satisfies \textbf{two} key properties: First, it must aggregate potentially weak and spatially distributed cues across the image rather than rely on a single dominant signal. This ensures that camouflage characterized by fragmented or subtle evidence is properly captured, avoiding single-point or single-region evaluation. Second, the score is conditioned on a known target category $c$. Conditioning constrains the search space to semantically relevant regions, reflecting the fact that camouflage difficulty is evaluated relative to a known target class. Together, these properties define camouflage as a category-aware, multi-cue localization problem.

As shown in Figure~\ref{fig:method_overview}, our multi-cue proposal aggregation framework proceeds in two stages (1) semantic proposal generation (2) camouflage cue integration.

\paragraph{\textbf{Stage I: Semantic Proposal Generation.}}

\begin{algorithm}[t]
\caption{\ourmethod: Multi-Cue Camouflage Scoring}
\label{alg:seamcam}
\small
\begin{algorithmic}[1]
\Require Image $I$, ground-truth masks $\mathcal{M} = \{M_j\}_{j=1}^{m}$, target category $c$
\Require Detector $\mathcal{D}$, segmentor $\mathcal{S}$, thresholds $\tau_\alpha, \tau_\beta$, maximum proposals $K_{\max}$
\Ensure Camouflage score $\zeta(I,\mathcal{M},c) \in [0,1]$
\vspace{1mm}
\State $\{(B_i,\alpha_i,\beta_i)\}_{i=1}^{N} \gets \mathcal{D}(I,c)$ \Comment{Generate semantic proposals}
\State $\mathcal{P} \gets \emptyset$
\For{$i=1,\ldots,N$}
    \If{$\alpha_i \geq \tau_\alpha$ \textbf{and} $\beta_i \geq \tau_\beta$}
        \State $\mathcal{P} \gets \mathcal{P} \cup \{B_i\}$
    \EndIf
\EndFor
\State $K \gets \min\{K_{\max}, |\mathcal{P}|\}$
\State $\mathcal{P} \gets \text{Top-}K(\mathcal{P})$ \Comment{Ranked by detection confidence}
\vspace{1mm}
\For{$i=1,\ldots,K$}
    \State $\widehat{M}_i \gets \mathcal{S}(I; B_i)$ \Comment{Segment each proposal}
\EndFor
\vspace{1mm}
\State $M^* \gets \bigcup_{j=1}^{m} M_j$ \Comment{Ground-truth union mask}
\State $D \gets 0$
\For{each non-empty subset $S \subseteq \{1,\ldots,K\}$}
    \State $\widehat{M}(S) \gets \bigcup_{i \in S} \widehat{M}_i$
    \State $\phi(S) \gets \mathrm{IoU}(\widehat{M}(S), M^*)$
    \State $D \gets \max(D,\phi(S))$
\EndFor
\vspace{1mm}
\State \Return $\zeta(I,\mathcal{M},c) = 1 - D$
\end{algorithmic}
\end{algorithm}

We query the SOTA open-vocabulary detector model, Grounding DINO~\cite{liu2024grounding}, with image $I$ and animal category prompt $c$, producing candidate detections
\begin{equation}
    \mathcal{D}(I, c) \;\rightarrow\; \big\{(B_i, \alpha_i, \beta_i)\big\}_{i=1}^{N},
\end{equation}
where $B_i$ is a proposal region bounding box, $N$ is the number of  proposal bounding boxes, $\alpha_i \in [0,1]$ is the text--image alignment score, and $\beta_i \in [0,1]$ is the detection confidence.

We observe a natural consequence of camouflage where detection confidence becomes asymmetric: true positives often receive suppressed confidence due to visual concealment. To address this, we apply two complementary gating mechanisms that mitigate this issue. First, we enforce strong text alignment by thresholding $\alpha_i \geq \tau_\alpha$, thereby filtering semantically irrelevant detections. Second, we preserve nuanced and weak detections by using a confidence threshold, $\beta_i \geq \tau_\beta$ which allows subtle yet potentially informative cues to be retained. From set $\mathcal{D}$, we retain the top-$K$ proposals ranked by confidence, 
$K = \min\{K_{\max}, N\}$, yielding the final proposal set $\mathcal{P} = \{B_1, \ldots, B_K\}$. 
This selection ensures a reasonable tradeoff between latency and accuracy; based on our 
hyperparameter search (Appendix Section~\ref{sec:appendix:hyperparams}), we set top-$K$ to $7$, $\tau_\alpha$ 
to $0.50$, and $\tau_\beta$ to $0.10$.
 
\paragraph{\textbf{Stage II: Camouflage Cue Integration.}}

\begin{algorithm}[t]
\caption{\ourmethod DPO Training for Camouflage Generation}
\label{alg:dpo}
\small
\begin{algorithmic}[1]
\Require Training set $\{(I_i, \mathcal{M}_i, c_i)\}$, VLM $\mathcal{V}$, reference diffusion model $\pi_{\text{ref}}$, trainable model $\pi_\theta$
\Require Number of prompts $X$, \ourmethod scorer $\zeta$, DPO temperature $\beta$
\For{each training image $(I, \mathcal{M}, c)$}
    \State $F \gets \text{ExtractForeground}(I, \mathcal{M})$
    \State $y^w \gets I$ \Comment{Natural camouflage image (winner)}
    \State $\{p_1,\dots,p_X\} \gets \mathcal{V}(F)$ \Comment{Generate prompt variations}
    \For{$k = 1,\dots,X$}
        \State $y_k \gets \pi_{\text{ref}}(\cdot \mid F, p_k)$ \Comment{Generate candidate camouflage}
        \State $s_k \gets \zeta(y_k, \mathcal{M}, c)$ \Comment{Evaluate camouflage strength}
    \EndFor
    \State $y^l \gets y_{\arg\max_k s_k}$ \Comment{Select strongest camouflaged candidate}
    \State Store preference pair $(y^w \succ y^l)$
\EndFor
\State Optimize $\pi_\theta$ using DPO objective $\mathcal{L}_{\text{DPO}}$ on collected preference pairs
\end{algorithmic}
\end{algorithm}

For each proposal $B_i \in \mathcal{P}$, we obtain a segmentation mask using SAM-2~\cite{ravi2024sam} $\widehat{M}_i = \mathcal{S}(I; B_i)$. To model multi-cue integration, we evaluate all possible permutations of non-empty subsets of $S \subseteq \{1, \ldots, K\}$, yielding $2^K-1$ combinations, from which we form the union mask $\widehat{M}(S) = \bigcup_{i \in S} \widehat{M}_i$ and compare it against the ground-truth union $M^* = \bigcup_{j=1}^{m} M_j$.

We define the subset score as $\phi(S) = \mathrm{IoU}\big(\widehat{M}(S), M^*\big)$. Then, the detectability of a camouflaged target is defined as the best achievable overlap between any subset of proposal cues and the ground-truth object mask:
\begin{equation}
    D(I, \mathcal{M}, c) \;=\; \max_{\emptyset \neq S \subseteq \{1,\ldots,K\}} \phi(S).
\end{equation}

\textbf{Remark:} Intuitively, this formulation selects the subset of proposals whose combined evidence yields the highest agreement with the ground-truth object region. We refer to this quantity as the \emph{detectability} score $D$, which represents the maximum recoverable signal available to an observer given the available cues.

From detectability we derive the proposed camouflage metric, \textbf{\ourmethod}, defined as
\begin{equation}
    \mathrm{\zeta}(I,\mathcal{M},c) = 1 - D(I,\mathcal{M},c).
\end{equation}

The detectability score $D$ can be interpreted as an upper bound on how well a capable observer could recover the camouflaged object under ideal conditions: (i) the observer knows the object category, (ii) they are able to identify all relevant visual cues, and (iii) they optimally integrate these cues. High values of $D$ therefore indicate that the target remains readily discoverable, whereas low values imply that cue integration produces minimal overlap with the ground-truth object mask, making the camouflaged object difficult to locate. Consequently, lower detectability corresponds to stronger camouflage and yields a higher \ourmethod score. Algorithm~\ref{alg:seamcam} summarizes the complete procedure.

The maximization over subsets also provides two desirable robustness properties. First, duplicate or highly overlapping proposals—commonly produced during detector inference—do not affect the value of $D$, since redundant cues do not increase the maximum achievable overlap. Second, the maximization naturally emphasizes the strongest and most informative cues that reveal the camouflaged object, while ignoring weaker or noisy evidence that does not contribute to improved localization.

\subsection{\ourmethod Preference Learning for Camouflage Generation}

While modern diffusion models have demonstrated remarkable success in general-purpose image synthesis and inpainting~\cite{lugmayr2022repaint,li2023gligen}, camouflage generation poses a fundamentally different challenge. Image inpainting is typically formulated as a conditional image restoration problem in which the objective is to reconstruct missing or corrupted regions such that the completed image remains photorealistic and consistent with the surrounding context~\cite{ho2020denoising,songdenoising}. However, effective camouflage requires an additional property: the generated object must not only appear realistic, \textit{but also} remain difficult for observers to detect. Standard diffusion training does not explicitly account for this requirement, as its objectives focus on reconstructing image distributions or plausibly filling missing regions. Consequently, general-purpose inpainting models such as SD-V2~\cite{rombach2022high} may produce visually plausible compositions while failing to achieve the subtle foreground–background integration necessary for convincing camouflage. This observation motivates an important question: beyond evaluating camouflage quality, can \ourmethod be used to effectively improve camouflage generation ?

To explore this question, we introduce a preference-based training paradigm that incorporates \ourmethod directly into the learning process. Our approach builds upon recent advances in Direct Preference Optimization (DPO)~\cite{rafailov2023direct,wallace2024diffusion}, which enable generative models to learn from pairwise preference signals without requiring expensive reinforcement learning training. We propose that DPO provides an effective framework for leveraging \ourmethod as a training signal for camouflage generation. Specifically, for each training instance, we treat the real image from our training datasets, COD-10k~\cite{fan2020camouflaged} and CAMO-FS~\cite{nguyen2024art}, as the \emph{winner} sample representing an ideal camouflage configuration observed in nature. Candidate camouflage images are then generated using a reference diffusion model, and \ourmethod is used to score and rank these candidates according to their camouflage strength. The candidate exhibiting the strongest camouflage (highest \ourmethod score) is selected as the hard negative in the preference pair. By training the target diffusion model to prefer the natural image over this challenging alternative, the model learns subtle distinctions that simple reconstruction losses cannot capture, as also demonstrated in~\cite{na2025boost}. Through this \ourmethod-guided preference curriculum, the model gradually internalizes the perceptual cues underlying effective camouflage. As demonstrated in our experiments in section~\ref{exp:dpo_as_prefer}, incorporating \ourmethod into the training loop leads to substantially more realistic and harder-to-detect camouflage synthesis compared to baseline diffusion models.

\begin{figure*}[t]
    \centering
    \includegraphics[width=\textwidth]{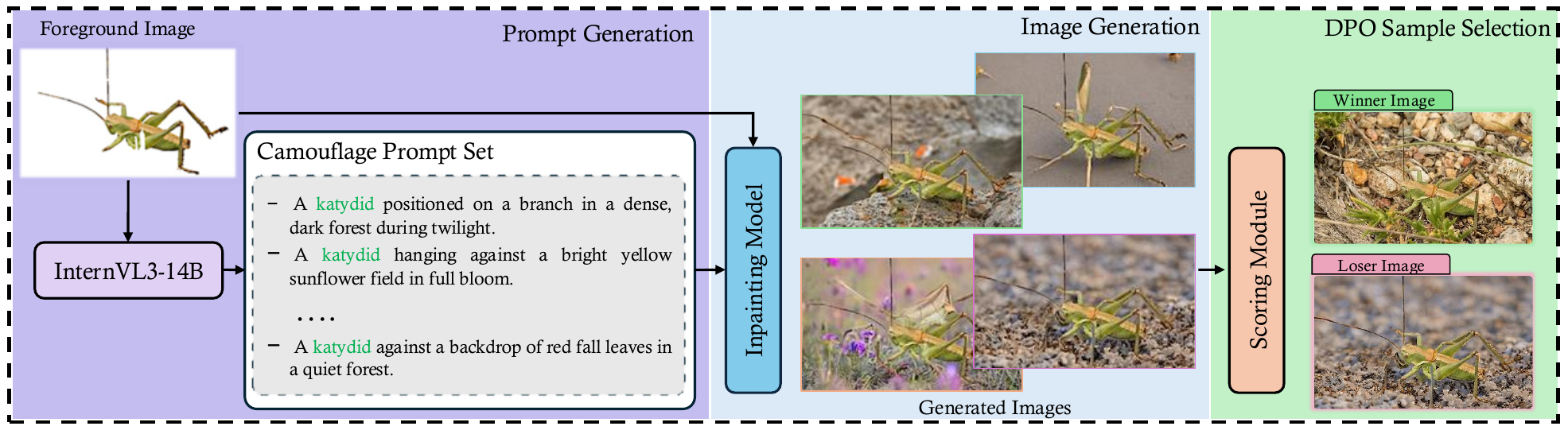}
    \vspace{-2mm}
    \caption{\textbf{\ourmethod-based sample selection for Direct Preference Optimization.} Using InternVL3-14B VLM, we generate diverse camouflage prompts describing different environmental contexts. An inpainting diffusion model produces multiple camouflaged candidate images conditioned on these prompts. Each candidate is given a score by the scorer, and the highest-scoring image is selected as the loser, while the winner image is picked from the dataset, forming a preference pair used for subsequent DPO training. The loser acts as the hard-negative in the DPO setting.}
    \label{fig:dpo_pipeline}
    \vspace{-3mm}
\end{figure*}

We now describe the \ourmethod-guided preference curriculum in detail. 
Let $(I, \mathcal{M}, c)$ denote a training sample drawn from the COD-10K and CAMO-FS datasets, where $I$ is the image, $\mathcal{M}$ is the foreground animal mask, and $c$ denotes the species category. The real image $I$ is treated as the preferred (winner) sample, reflecting naturally occurring camouflage that we aim to approximate during generation. Our training pipeline begins by extracting the foreground animal from the image using the provided mask. A vision–language model (InternVL3-14B) is then prompted to generate a set of 12 textual descriptions conditioned on the foreground subject, introducing variations in environmental context such as biome, lighting conditions, and seasonal appearance as shown in Figure~\ref{fig:dpo_pipeline}.

Next, we initialize two identical copies of a pretrained diffusion inpainting model (Stable Diffusion v2 Inpainting), denoted as the reference model $\pi_{\text{ref}}$ and the trainable model $\pi_\theta$. For each prompt generated by the vision–language model, the reference model $\pi_{\text{ref}}$ synthesizes a candidate camouflage image conditioned on the extracted foreground animal and the prompt description. This process produces a set of candidate generations that vary in camouflage quality. We evaluate each generated image using the proposed \ourmethod metric, which measures the perceptual difficulty of detecting the foreground object. The candidate with the highest \ourmethod score—corresponding to the most visually convincing camouflage among the generated samples—is selected as the hard negative example. This strategy follows the principle of \textbf{hard-negative mining}~\cite{wallace2024diffusion} in preference learning, where training focuses on challenging examples that are visually close to the desired outcome yet still imperfect. By forcing the model to distinguish between the natural camouflage image and a high-quality but synthetic alternative, the training process encourages the diffusion model to learn subtle visual cues that contribute to realistic concealment. The resulting preference pairs are used to optimize $\pi_\theta$ using the standard DPO objective, gradually guiding the generative process toward producing more convincing camouflage patterns. The full algorithmic procedure is outlined in Algorithm~\ref{alg:dpo}.

\section{Experiments}
\label{sec:experiments}

We evaluate \ourmethod along two complementary dimensions corresponding to the primary contributions of this work. First, we assess \ourmethod as a camouflage evaluation metric by measuring its alignment with human perceptual judgments. Second, we evaluate \ourmethod as a training signal for preference-based optimization by using it to guide Direct Preference Optimization (DPO) fine-tuning of a SD-V2 inpainting diffusion model~\cite{rombach2022high}. The former experiment examines whether \ourmethod captures perceptual camouflage difficulty in a manner consistent with human observers, while the latter tests whether optimizing for \ourmethod leads to the synthesis of more effectively camouflaged images. Further details on datasets, hyperparameter selection, and DPO training are provided in Appendix~\ref{sec:appendix:datasets},~\ref{sec:appendix:hyperparams}, and~\ref{sec:appendix:training}, respectively.

\begin{figure*}[ht]
    \centering
    \includegraphics[width=\linewidth]{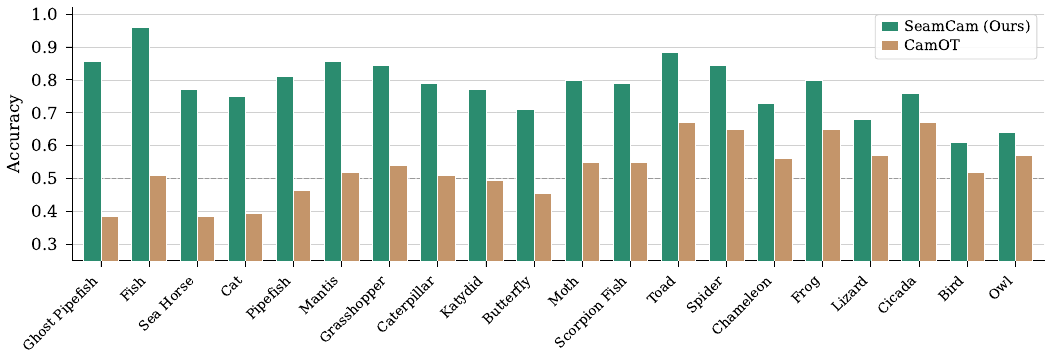}
    \vspace{-8mm}
    \caption{\textbf{Per-species accuracy comparison between \ourmethod vs. CamOT}. In our 2AFC human study, \ourmethod (green) consistently outperforms CamOT (brown) across all categories, with an average accuracy of $\sim$0.79 compared to $\sim$0.53 for CamOT. The largest performance gaps are observed in \textit{Fish} ($\Delta = 0.45$), \textit{Sea Horse} ($\Delta = 0.39$), and \textit{Cat} ($\Delta = 0.36$), while the smallest margins occur in \textit{Cicada} and \textit{Toad}. The horizontal line denotes the 0.5 chance-level baseline.}
    \label{fig:forest-per-species} 
\end{figure*}

\subsection{Evaluation of \ourmethod as a Camouflage Metric}

We first evaluate whether \ourmethod aligns with human perception of camouflage difficulty. The central question of this study is whether \ourmethod can judge the relative difficulty of detecting a camouflaged animal in a manner consistent with human observers, and how this alignment compares to CamOT~\cite{das2025camouflage}.

We adopt a Two-Alternative Forced Choice (2AFC) protocol~\cite{ulrich2004threshold}, as humans are generally more reliable at comparative judgments than absolute scoring. Participants are presented with pairs of images and asked to select the image in which the animal is harder to find. The majority vote for each pair is treated as the human ground truth. For each image pair, \ourmethod and CamOT independently assign camouflage scores to both images. The metric predicts the “harder” image by selecting the one with the higher camouflage score. Performance is quantified as the proportion of image pairs for which the metric’s prediction agrees with the human majority, yielding an accuracy measure of perceptual alignment. In total, we collected 2,390 pairwise judgments from 94 participants. \ourmethod achieves an agreement accuracy of \textbf{78.82\%}, while CamOT achieves \textbf{54.07\%}. Notably, CamOT performs only marginally above random guessing, indicating limited alignment with human perceptual judgment. In contrast, \ourmethod demonstrates substantially stronger agreement, supporting its validity as a perceptually grounded camouflage metric. Figure~\ref{fig:forest-per-species} shows the mean accuracy performance of \ourmethod vs. CamOT in our human study.

\begin{table*}[t]
\centering
\small
\caption{\textbf{Ablation study on training signal and preference pair selection using SD-V2 as the base model}. DPO-based fine-tuning consistently outperforms zero-shot and LoRA + SFT baselines. Among DPO variants, SeamCam-based preference selection achieves the best SeamCam score and the best KID and FID, while DPO (CamOT)'s inflated CamOT score is attributed to in-distribution bias, revealed when compared to the \ourmethod score. Bold indicates best, underline indicates second best.}
\resizebox{\textwidth}{!}{
\begin{tabular}{l | c | c c | c | c c}
\toprule
\rowcolor{gray!10}
\textbf{Base} & \textbf{Training Signal}
& \textbf{KID}$\downarrow$ & \textbf{FID}$\downarrow$
& \textbf{HPS-v2}$\uparrow$
& \textbf{CamOT}$\uparrow$ & \textbf{\ourmethod}$\uparrow$ \\
\midrule
\multirow{6}{*}{SD-V2} & Zero Shot                  & 0.0258 & 88.28 & 0.190 & 0.3619 & 0.1824 \\
                       & LoRA + SFT         & 0.0248 & 84.28 & \underline{0.193} & 0.3641 & 0.2519 \\
\cmidrule(lr){2-7}
                       & DPO (Random)       & 0.0187 & 78.04 & 0.192 & 0.3874 & 0.2565 \\
                       & DPO (CamOT)        & \underline{0.0168} & \underline{77.26} & \underline{0.193} & \textbf{0.4352} & \underline{0.2995} \\
                       & \textbf{DPO (\ourmethod)} & \textbf{0.0139} & \textbf{71.25} & \textbf{0.195} & \underline{0.4144} & \textbf{0.3811} \\
\bottomrule
\end{tabular}
}
\label{tab:dpo_ablation}
\vspace{-4mm}
\end{table*}

\begin{table}[t]
\centering
\small
\caption{\small \textbf{Comparison with camouflage generation methods on \ourdataset}. DPO (\ourmethod) is evaluated against prior inpainting and generation baselines across distributional realism (KID, FID), human preference alignment (HPS-v2), and camouflage quality (CamOT, \ourmethod). Our model achieves the best KID, HPS-v2, and \ourmethod scores, confirming that \ourmethod-guided preference optimization yields both higher image fidelity and more effective camouflage generation. Bold indicates best, underline indicates second best.}
\begin{tabular}{l|cc|c|cc}
\toprule
\rowcolor{gray!10}
\textbf{Method} & \textbf{KID}$\downarrow$ & \textbf{FID}$\downarrow$ & \textbf{HPS-v2}$\uparrow$ & \textbf{CamOT}$\uparrow$ & \textbf{\ourmethod}$\uparrow$ \\
\midrule
LCG-Net~\cite{li2022location} & 0.1455 & 248.58 & 0.158 & \underline{0.4263} & 0.2954 \\
TFill~\cite{zheng2022bridging} & 0.0669 & 131.30 & 0.148 & \textbf{0.5067} & 0.3571 \\
LDM~\cite{rombach2022high} & 0.0423 & 99.43 & 0.165 & 0.3751 & 0.3474 \\
RePaint-L~\cite{lugmayr2022repaint} & 0.0266 & 101.21 & \underline{0.172} & 0.3930 & 0.3501 \\
LAKE-RED~\cite{zhao2024lake} & \underline{0.0157} & \textbf{70.65} & 0.171 & 0.3967 & \underline{0.3639} \\
\midrule
\textbf{DPO (\ourmethod)} & \textbf{0.0139} & \underline{71.25} & \textbf{0.195}& 0.4144 & \textbf{0.3811} \\
\bottomrule
\end{tabular}
\vspace{1mm}
\label{tab:gen_vs_sota}
\vspace{-4mm}
\end{table}

\begin{figure}[ht]
    \centering
    \includegraphics[width=\linewidth]{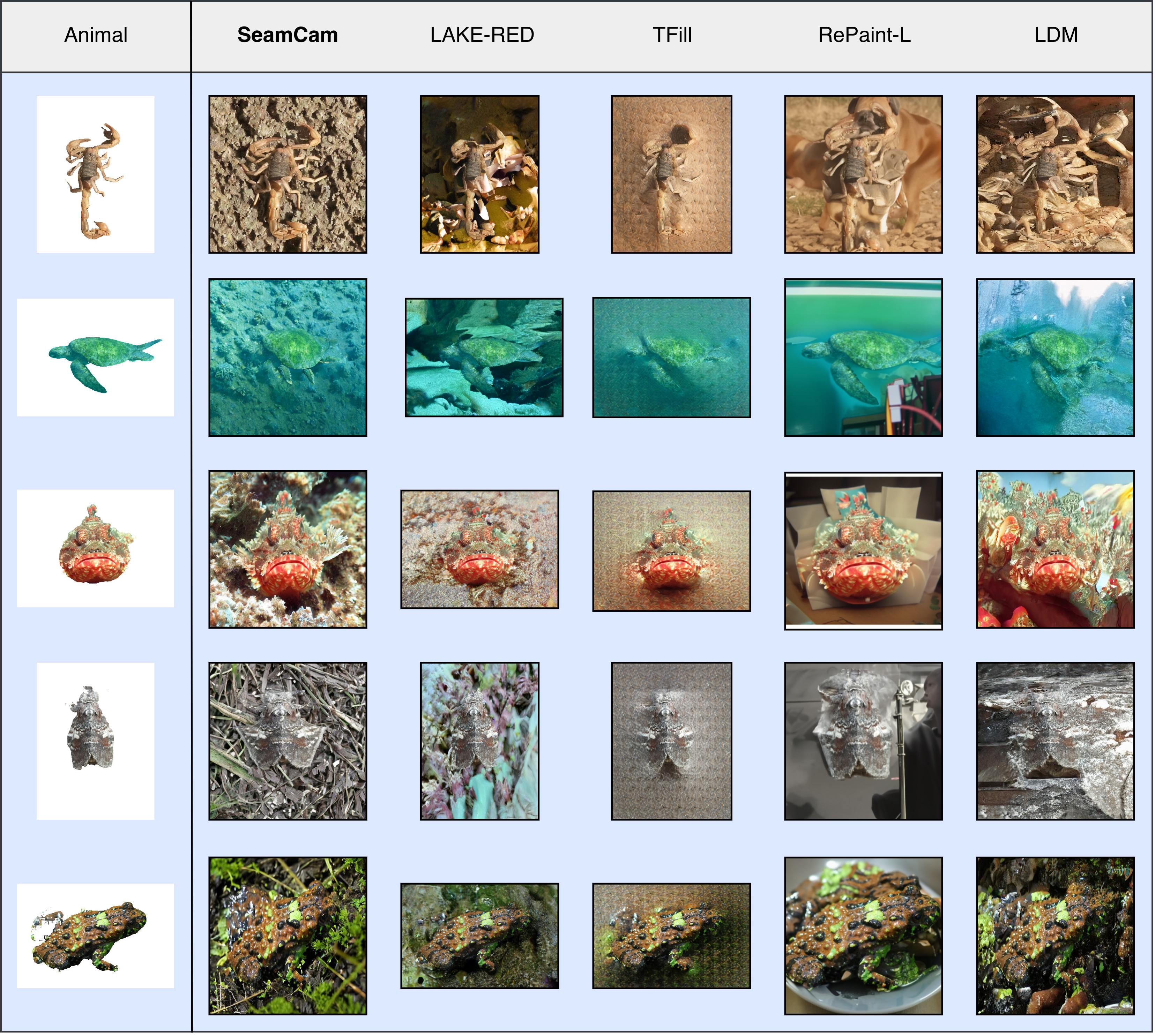}
    \vspace{-7mm}
    \caption{\small \textbf{Camouflage image generation using \ourmethod-based DPO \emph{vs.} baselines.}
    Given only the foreground mask (left) from \ourdataset, \ourmethod produces visually coherent backgrounds that are semantically appropriate to each animal's natural habitat. Baselines fail in distinct ways: TFill loses color fidelity, RePaint-L generates implausible scenes, LDM produces over-saturated artifacts, and LAKE-RED places animals in ecologically mismatched environments despite appearing superficially plausible.
    }
    \label{fig:camgen_sota}
    \vspace{-4mm}
\end{figure}

\subsection{\ourmethod as a Preference Signal for DPO Training}
\label{exp:dpo_as_prefer}

We next investigate \ourmethod as a training signal for camouflage synthesis, using it for hard-negative selection during Direct Preference Optimization (DPO) fine-tuning of a diffusion-based inpainting model. To isolate the contribution of the scoring function, we fix the base architecture to SD-V2 and vary only the training objective and preference pair selection strategy across five settings. Table~\ref{tab:dpo_ablation} presents two key findings. First, DPO-based fine-tuning consistently outperforms both the zero-shot baseline and LoRA + SFT fine-tuning, demonstrating that preference optimization is a more effective training signal for camouflage synthesis than supervised approaches. Second, the choice of preference pair selection criterion matters: \ourmethod-based selection outperforms both random and CamOT-based selection. Crucially, while DPO (CamOT) achieves the highest CamOT score, this is attributable to in-distribution reward hacking rather than genuine camouflage quality improvement — the model simply overfits to the metric used to curate its own training pairs. DPO (\ourmethod), by contrast, achieves superior performance on the scorer-independent metrics (KID and FID), confirming that \ourmethod provides a more faithful training signal.

Having established \ourmethod-guided DPO as the strongest training configuration, we compare it against prior camouflage generation methods in Table~\ref{tab:gen_vs_sota}. Using the curated CamFG-1.5k dataset, we benchmark our DPO-trained inpainting model guided by \ourmethod against five baseline methods: LCG-Net~\cite{li2022location}, TFILL~\cite{zheng2022bridging}, LDM~\cite{rombach2022high}, RePaint-L~\cite{lugmayr2022repaint}, and LAKE-RED~\cite{zhao2024lake}. We report both generative quality metrics FID~\cite{heusel2017gans} and KID~\cite{binkowski2018demystifying}, HPS-v2~\cite{wu2023human} and camouflage-specific metrics CamOT~\cite{das2025camouflage} and \ourmethod scores. Our model achieves the best performance across image plausibility, quality and camouflage effectiveness simultaneously. Additionally, Figure~\ref{fig:camgen_sota} demonstrates these results qualitatively.

\section{Discussion and Conclusion}
\label{sec:conclusion}

We introduced \textbf{\ourmethod}, a perceptually grounded metric for quantifying camouflage effectiveness by reframing camouflage evaluation as a visual localization problem. Rather than measuring foreground–background similarity, SeamCam estimates the maximum recoverable localization signal from category-conditioned detection proposals, directly capturing how detectable an animal remains given available visual evidence. Through a large-scale two-alternative forced-choice study with 94 participants and 2,390 comparisons, we demonstrated that SeamCam aligns strongly with human judgments of camouflage difficulty, substantially outperforming prior metrics. Beyond evaluation, we showed that SeamCam provides a reliable preference signal for Direct Preference Optimization, enabling diffusion-based inpainting models to generate images that are both more difficult to detect and higher in visual fidelity, while avoiding reward-hacking behavior observed with alternative metrics. To support rigorous and unbiased benchmarking, we introduced \textbf{\ourdataset}, a curated dataset of fully visible animals that isolates camouflage synthesis from occlusion artifacts present in existing evaluation datasets. Together, \ourmethod and \ourdataset establish a principled framework for evaluating and optimizing camouflage generation in a manner consistent with human perceptual judgment.

\bibliographystyle{splncs04}
\bibliography{main}

\clearpage
\appendix
\setcounter{figure}{0}
\setcounter{table}{0}
\renewcommand{\thetable}{S\arabic{table}}  
\renewcommand{\thefigure}{S\arabic{figure}}
\section*{Appendix}




\label{sec:appendix}

This appendix provides supplementary details that support the main paper.
Section~\ref{sec:appendix:datasets} describes the training and evaluation datasets,
including the curation procedure for \textbf{\ourdataset}.
Section~\ref{sec:appendix:hyperparams} details the staged hyperparameter search used
to configure \ourmethod. Section~\ref{sec:appendix:training} provides full details of
the DPO training procedure, including the prompt design for both the inpainting model
and the vision--language model used for preference pair construction.
Section~\ref{sec:appendix:human_study} provides the full protocol and statistical
analysis of the 2AFC human study. Section~\ref{sec:appendix:backbone}
analyzes the robustness of \ourmethod across different detector/segmenter backbones.
Section~\ref{sec:appendix:disruptive} examines \ourmethod's behavior on disruptive
coloration and other special camouflage strategies.
Section~\ref{sec:appendix:biome} reports the biome diversity of the evaluation set.
Section~\ref{sec:appendix:detector_behavior} analyzes detector confidence under
occlusion versus camouflage.
Section~\ref{sec:appendix:runtime} provides a detailed runtime breakdown.
Section~\ref{sec:appendix:cod} reports zero-shot COD evaluation on DPO-generated outputs
and discusses additional downstream uses for camouflage detection.

\section{Dataset Details}
\label{sec:appendix:datasets}

\subsection{Training Datasets}
\label{sec:appendix:train_dataset}

For training the DPO-based inpainting model, we use COD-10k~\cite{fan2020camouflaged}
and CAMO-FS~\cite{nguyen2024art}, which together provide a diverse range of species
and natural camouflage scenarios. COD-10k contains 10,000 images spanning 78 categories,
while CAMO-FS includes 1,250 images across 8 superclasses. These datasets offer varied
ecological settings and camouflage patterns suitable for learning realistic concealment
strategies.

\subsection{CamFG-1.5k: Evaluation Dataset}
\label{sec:appendix:eval_dataset}

A critical limitation of existing camouflage datasets is that many images contain
animals that are already partially camouflaged, cropped, or significantly occluded ---
making it difficult to isolate and objectively assess the quality of generated camouflage.
To address this, we introduce \textbf{CamFG-1.5k}, a carefully curated benchmark of 1{,}521
high-resolution images sourced from iNaturalist~\cite{van2018inaturalist}. Each image
contains a complete, well-segmented animal with minimal occlusion, enabling rigorous
evaluation of background synthesis and camouflage generation from fully visible subjects.
The dataset spans more than 1{,}000 species across six taxonomic groups (insects, reptiles,
amphibians, fish, birds, and mammals), covers multiple biome types, and includes varied
lighting conditions. CamFG-1.5k will be publicly released to support reproducible and
unbiased benchmarking of camouflage generation methods.

\paragraph{\textbf{Disjointness from Training Sources.}}
\ourdataset\ is curated entirely from iNaturalist, an image source that is separate from
both COD-10k and CAMO-FS. Disjointness between training and evaluation therefore holds
\emph{by construction} rather than by post-hoc filtering, making \ourdataset\ a genuine
out-of-distribution evaluation set for our DPO-trained model. While species-level overlap
across general camouflage benchmarks is unavoidable and follows standard practice in the
field, image-level separation is fully maintained.

\section{\ourmethod Hyperparameter Selection}
\label{sec:appendix:hyperparams}

\pgfplotstableread{
x y val
0 0 78.82
1 0 78.82
2 0 78.82
3 0 71.57
4 0 63.69
5 0 56.99
6 0 52.95
7 0 51.33
8 0 50.48
0 1 78.82
1 1 78.82
2 1 78.82
3 1 71.57
4 1 63.69
5 1 56.99
6 1 52.95
7 1 51.33
8 1 50.48
0 2 78.82
1 2 78.82
2 2 78.82
3 2 71.57
4 2 63.69
5 2 56.99
6 2 52.95
7 2 51.33
8 2 50.48
0 3 78.82
1 3 78.82
2 3 78.82
3 3 71.57
4 3 63.69
5 3 56.99
6 3 52.95
7 3 51.33
8 3 50.48
0 4 74.10
1 4 74.10
2 4 74.10
3 4 67.85
4 4 60.00
5 4 53.30
6 4 49.26
7 4 47.64
8 4 46.79
0 5 64.65
1 5 64.65
2 5 64.65
3 5 58.40
4 5 50.56
5 5 43.86
6 5 39.82
7 5 38.20
8 5 37.35
0 6 57.96
1 6 57.96
2 6 57.96
3 6 51.71
4 6 43.86
5 6 37.16
6 6 33.12
7 6 31.50
8 6 30.65
0 7 54.11
1 7 54.11
2 7 54.11
3 7 47.86
4 7 40.01
5 7 33.31
6 7 29.27
7 7 27.65
8 7 26.80
0 8 51.49
1 8 51.49
2 8 51.49
3 8 45.24
4 8 37.39
5 8 30.69
6 8 26.65
7 8 25.03
8 8 24.18
}\heatmapdata

\begin{figure}[ht]
\centering
\resizebox{\textwidth}{!}{%
\begin{tikzpicture}

\begin{axis}[
    name=plot1left,
    axis y line*=left,
    axis x line=bottom,
    xlabel={Top-$K$},
    ylabel={Accuracy (\%)},
    ylabel style={color=acccolor},
    yticklabel style={color=acccolor},
    ytick style={color=acccolor},
    xmin=0.5, xmax=12.5,
    ymin=71, ymax=80,
    xtick={1,3,5,7,10,12},
    ymajorgrids=true,
    grid style={dotted, gray!40},
    tick align=outside,
    title={\textbf{(a) Top-$K$ Performance}},
    width=8.5cm, height=6.5cm,
]
    \addplot[color=acccolor, thick, mark=*]
    coordinates {(1,73)(3,75)(5,75)(7,78.1)(10,78.151)(12,78.159)};
    \addplot[only marks, mark=*, mark size=4pt,
    mark options={fill=acccolor!50, draw=acccolor, thick}]
    coordinates {(7,78.1)};
\end{axis}

\begin{axis}[
    at={(plot1left.south west)},
    anchor=south west,
    axis y line*=right,
    axis x line=none,
    ylabel={Latency (s)},
    ylabel style={color=latcolor},
    yticklabel style={color=latcolor},
    ytick style={color=latcolor},
    xmin=0.5, xmax=12.5,
    ymin=0.50, ymax=0.62,
    width=8.5cm, height=6.5cm,
]
    \addplot[color=latcolor, thick, mark=square*, dashed]
    coordinates {(1,0.556)(3,0.560)(5,0.562)(7,0.565)(10,0.565)(12,0.565)};
\end{axis}

\begin{axis}[
    at={(plot1left.outer north east)},
    anchor=outer north west,
    xshift=1.5cm,
    width=8.5cm, height=8.5cm,
    axis equal image,
    enlargelimits=false,
    xlabel={Box Threshold},
    ylabel={Text Threshold},
    xtick={0,...,8},
    xticklabels={0.01,0.05,0.10,0.15,0.20,0.25,0.30,0.35,0.40},
    ytick={0,...,8},
    yticklabels={0.05,0.15,0.30,0.50,0.70,0.80,0.85,0.90,0.95},
    xticklabel style={rotate=45, anchor=east},
    colormap/viridis,
    colorbar,
    point meta min=20, point meta max=80,
    colorbar style={title={Acc (\%)}},
    title={(b) \textbf{Threshold Grid Search}},
]
\addplot[
    matrix plot*,
    mesh/cols=9,
    mesh/rows=9,
    point meta=explicit,
] table [meta=val] {\heatmapdata};
    \draw[draw=red, thick] (axis cs:1.5,2.5) rectangle (axis cs:2.5,3.5);
\end{axis}
\end{tikzpicture}
}
\caption{\textbf{\ourmethod Hyperparameter Selection.} Results of the staged search
performed on 100 random samples from the evaluation set. \textbf{(a)} Accuracy--latency
trade-off across Top-$K$ values; $K{=}7$ is selected as the optimal elbow point.
\textbf{(b)} Grid search heatmap over box threshold $\tau_\beta$ and text threshold
$\tau_\alpha$, with the selected combination ($\tau_\alpha{=}0.50$, $\tau_\beta{=}0.10$)
highlighted in red.}
\label{fig:seamcam_hyperparam}
\end{figure}
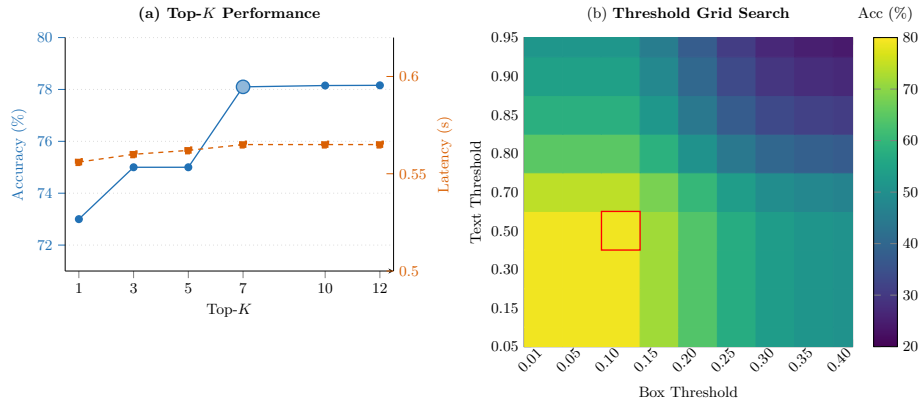

We perform a staged hyperparameter search on 100 randomly sampled pairs from the 2{,}390
pairwise comparisons collected in our human preference study. To decouple the search, we
first fix Top-$K{=}12$ and perform a grid search over the text threshold $\tau_\alpha$
and box threshold $\tau_\beta$, whose interaction is visualized in Figure~\ref{fig:seamcam_hyperparam}(b).
The optimal combination $\tau_\alpha{=}0.50$ and $\tau_\beta{=}0.10$ (highlighted in red)
yields the highest agreement with human judgments. With these thresholds fixed, we then
sweep over Top-$K$ values and identify $K{=}7$ as the elbow point that best balances
accuracy and latency, as shown in Figure~\ref{fig:seamcam_hyperparam}(a). All subsequent
experiments use these values.

\section{DPO Training Details}
\label{sec:appendix:training}

We provide full implementation details of the \ourmethod-guided preference learning
framework introduced in Section~\ref{exp:dpo_as_prefer}. We first describe the prompt
design strategy for both the inpainting model and the vision--language model, then
detail the optimization setup used for DPO fine-tuning.

\subsection{Prompt Design}
\label{sec:appendix:prompts}

\paragraph{\textbf{Inpainting Model Prompt.}} During both DPO training and inference of the SD-V2 inpainting model, we fix the conditioning text to the simple instruction:
\[
\text{text} = \text{``A camouflaged image of } \{c\}\text{''}
\]
This deliberately minimal prompt avoids overfitting to specific environmental descriptions, encouraging the model to learn camouflage from preference signal rather than from prompt-driven scene specification.

\paragraph{\textbf{VLM Background Generation Prompt.}} To construct preference pairs for DPO training, we extract the foreground animal from each training image using its provided mask and prompt InternVL3-14B to generate 12 candidate background descriptions conditioned on the foreground subject. These descriptions introduce controlled variation in environmental context --- biome, lighting conditions, and seasonal appearance --- which are then used to condition the reference inpainting model to synthesize a diverse pool of candidate camouflage images. The candidate with the highest \ourmethod score is selected as the hard negative in the preference pair. The prompt passed to the VLM is:

\vspace{1mm}
\begin{tcolorbox}[colback=blue!3, colframe=blue!40, boxrule=0.4pt, arc=2pt, left=2pt, right=2pt, top=2pt, bottom=2pt]
\small
\texttt{Given [SUBJECT], generate 12 prompts for natural backgrounds that would camouflage it. Vary biome, lighting, season. Specify what to avoid: contrasting colors, clutter near subject.}
\end{tcolorbox}
\vspace{1mm}

The full instruction is:

\begin{quote}\small
\emph{``Given \texttt{SUBJECT}, write 12 short/medium prompts that keep the \texttt{SUBJECT} unchanged and replace only the background with natural, photoreal scenes that strongly camouflage the subject (complementary colors or light/dark, shallow depth of field, minimal clutter). Vary biome/season/time of day. Return JSON \{\texttt{prompt}, \texttt{negative}\} where \texttt{negative} suppresses camouflage tones, clutter, extra animals, motion blur, and anything touching the subject.''}
\end{quote}

\subsection{Optimization Setup}
\label{sec:appendix:optim}

We fine-tune the SD-V2 Inpainting model with the DPO objective, updating only background pixels while keeping the foreground fixed. Following~\cite{wallace2024diffusion}, optimization uses AdamW with an effective batch size of 2{,}048 pairs on 16$\times$A100 GPUs (local batch size of 1 pair, gradient accumulation of 128 steps), at fixed square input resolution. The learning rate follows a $\beta$-aware scaling rule with learning rate of $1 \times 10^{-6}$,
\[
\eta(\beta) = \left(\frac{2000}{\beta}\right) \times 2.048 \times 10^{-6},
\]
with a 25\% linear warmup. The inverse dependence on $\beta$ stabilizes gradient step sizes, as the DPO gradient norm grows approximately linearly with $\beta$~\cite{rafailov2023direct}. For SD-V2 Inpainting we set $\beta{=}5000$, which we found to give strong convergence and stable updates.

\section{2AFC Human Study Protocol and Statistical Analysis}
\label{sec:appendix:human_study}

This section details the protocol used to collect the 2{,}390 pairwise judgments
reported in Section~\ref{sec:experiments}, together with additional statistical
analyses that further support the perceptual alignment of \ourmethod.

\subsection{Protocol}
\label{sec:appendix:human_study:protocol}

\paragraph{\textbf{Participants.}} 94 participants completed the study. Each participant
answered a fixed-size set of pairwise comparisons drawn from the species pool described
in the main paper. To improve response quality, participants were screened with
attention-check catch trials in which one of the two images was clearly non-camouflaged
(e.g., a high-contrast subject on a uniform background). Participants failing the catch
trials were excluded from the final analysis.

\paragraph{\textbf{Stimulus and Instruction.}} On each trial, two images of camouflaged
animals from the same species pool were displayed side-by-side. The instruction was
phrased as: \emph{``Which image is more camouflaged?''}. Image order (left/right) was
randomized per trial to remove side bias. We did not provide a time limit, but
participants were encouraged to answer based on their first perceptual impression.

\paragraph{\textbf{Aggregation Rule.}} For each image pair, the human ground-truth label
was defined as the majority vote across participants. Pairs in which a tie occurred or
fewer than three valid responses were collected were excluded from the analysis. After
these filters, 2{,}290 pairs remained with valid scores from both \ourmethod\ and CamOT;
all reported alignment figures use this shared evaluable subset.

\paragraph{\textbf{Metric-Side Evaluation.}} For each pair, each metric independently
assigns a score to both images and predicts the ``harder'' one as the higher-scoring
image. Accuracy is the proportion of pairs in which the metric prediction matches the
human majority. This evaluation is invariant under any monotone rescaling of metric
scores, and depends only on within-pair ordering.

\subsection{Statistical Analysis}
\label{sec:appendix:human_study:stats}

Beyond the headline agreement rates of \textbf{78.82\%} (\ourmethod) and \textbf{53.89\%}
(CamOT), we report additional statistical tests that confirm the gap is highly significant
and consistent across conditions.

\paragraph{\textbf{Paired McNemar Test.}} On the shared paired-evaluable subset
($n{=}2{,}271$, after removing per-method undecidable cases), the contingency for
``which metric agreed with humans'' yields $n_{01}{=}833$ pairs where \ourmethod\ alone
agreed and $n_{10}{=}259$ pairs where CamOT alone agreed. McNemar's chi-squared statistic
with continuity correction is
\[
\chi^2_{\mathrm{cc}} = \frac{(|n_{01}-n_{10}|-1)^2}{n_{01}+n_{10}} = 300.67,
\]
with $p < 10^{-67}$, corresponding to roughly a $3.2\times$ paired advantage for
\ourmethod. The result is highly significant under any reasonable correction.

\paragraph{\textbf{Bootstrap Confidence Intervals.}} Bootstrap 95\% confidence intervals
over pairs are non-overlapping between the two metrics, providing a distribution-free
confirmation of the gap.

\paragraph{\textbf{Per-Species Wilson Intervals.}} Computing per-species Wilson 95\%
confidence intervals (Fig.~5 of the main paper) places \ourmethod\ above the 0.5
chance-level baseline for all 20 categories, including the smallest-margin categories
(\textit{Cicada}, \textit{Toad}). In contrast, several CamOT per-species intervals
straddle or fall below chance.

\paragraph{\textbf{Score Gap vs.\ Human Consensus.}} To verify that metric ordering is
stable across pairs of varying human agreement, we computed the Spearman rank correlation
between the absolute metric score gap and the human vote margin. \ourmethod\ yields
$\rho{=}+0.011$, indicating ordering is essentially invariant to consensus level. CamOT
yields $\rho{=}-0.048$, indicating a slight inversion --- larger CamOT score gaps are, if
anything, weakly associated with smaller human consensus. This further supports
\ourmethod's perceptual alignment under both easy and ambiguous pairs.

\section{Backbone Robustness Study}
\label{sec:appendix:backbone}

\ourmethod\ is intentionally formulated as a \emph{framework}: any category-aware
detector and segmenter can be substituted into the pipeline (Algorithm~1). Because
camouflage is defined relative to an observer --- an animal is well-camouflaged
\emph{because} an observer cannot localize it --- the metric is naturally model-relative,
and tracks the perceptual capability of the underlying backbone as it improves. This
section directly tests how sensitive the metric is to the choice of backbone.

\paragraph{\textbf{Setup.}} We re-run \ourmethod\ on the 2AFC evaluation set under four
different detector/segmenter configurations: (i) our default Grounding
DINO~\cite{liu2024grounding} + SAM-2~\cite{ravi2024sam}; (ii) SAM-3, which performs
unified open-vocabulary detection and segmentation; (iii) Rex-Omni~\cite{jiang2025detect} + SAM-2; and
(iv) OWLv2-large~\cite{minderer2023scaling} + SAM-2. Hyperparameters ($\tau_\alpha{=}0.50$, $\tau_\beta{=}0.10$,
Top-$K{=}7$) are held fixed across configurations.

\begin{table}[ht]
\centering
\small
\setlength{\tabcolsep}{6pt}
\caption{\textbf{Backbone robustness of \ourmethod.} 2AFC agreement with human judgments
across four detector/segmenter configurations on the shared evaluable subset
($n{=}2{,}290$). All four pipelines stay in a tight $\sim$2-point band and preserve a
$\sim$24-point margin over CamOT (53.89\%).}
\label{tab:appendix:backbone}
\begin{tabular}{lcc}
\toprule
Pipeline & Accuracy (\%) & $\Delta$ vs.\ CamOT \\
\midrule
Grounding DINO + SAM-2 (ours) & 78.82 & +24.93 \\
SAM-3 (unified)               & 77.48 & +23.59 \\
Rex-Omni + SAM-2              & 79.69 & +25.80 \\
OWLv2-L + SAM-2               & 79.17 & +25.28 \\
\bottomrule
\end{tabular}
\end{table}

\paragraph{\textbf{Findings.}} As shown in Table~\ref{tab:appendix:backbone},
\ourmethod's agreement with human judgment remains in a tight \textbf{77.48--79.69\%}
band, preserving a $\sim$24-point margin over CamOT in every configuration. This indicates
that the metric's rankings track camouflage structure rather than artifacts of any
specific detector or segmenter.

\paragraph{\textbf{Why It Is Stable.}} Three properties of the design absorb
backbone-level variability: (i) \emph{semantic gating} ($\tau_\alpha$) removes proposals
that are not category-consistent; (ii) \emph{permissive confidence gating}
($\tau_\beta{=}0.10$) preserves the weak-but-valid cues that are characteristic of
camouflaged targets, which would otherwise be lost to standard high-confidence detection
thresholds; and (iii) \emph{subset-max aggregation} means that the score depends on
whether \emph{any} subset of proposals recovers the ground-truth extent rather than on
the success of a single detection. Together, these properties insulate the score from
proposal-level noise and produce the observed cross-backbone stability.

\paragraph{\textbf{Per-Species Consistency.}} Examining per-species performance under each
backbone, no category shows backbone-specific failure: across the 20 species in
Figure~5 of the main paper, every species remains above the 0.5 chance baseline for all
four pipelines. This is also relevant to the question of generalization to unseen species:
if the metric were dominated by backbone-specific biases, we would expect concentrated
degradation on a subset of taxa; instead, the gap to CamOT is broadly distributed across
categories.

\section{Disruptive Coloration and Special Camouflage Strategies}
\label{sec:appendix:disruptive}

A natural concern with any localization-based camouflage metric is whether it correctly
handles \emph{disruptive coloration} --- a biologically common strategy in which
high-contrast internal patterns break up the perceived outline of an animal rather than
matching its background. This section clarifies why \ourmethod\ explicitly rewards
disruptive coloration as a form of effective camouflage, in contrast to similarity-based
metrics such as CamOT which can misinterpret internal contrast as a lack of camouflage.

\paragraph{\textbf{Why \ourmethod\ Handles Disruptive Coloration Correctly.}}
\ourmethod\ does not take color similarity, edge smoothness, or any foreground--background
appearance distance as input. By definition (Eq.~3 in the main paper),
$\zeta = 1 - D$ where $D$ is the maximum IoU between the union of any subset of
category-conditioned proposals and the ground-truth mask. The score therefore depends on
\emph{localization failure}, not on appearance statistics. Disruptive coloration, by
fragmenting the perceived outline of the animal, causes detection proposals to either
miss the animal entirely or drift onto background regions sharing the disruptive
pattern. The resulting subset union fails to recover the ground-truth extent,
$D$ decreases, and $\zeta$ correctly rises.

This is the opposite behavior to CamOT, which compares foreground and background feature
distributions: high-contrast internal patterns increase the foreground--background
distance and are penalized as ``non-camouflage'' despite being highly effective at
preventing observer localization.

\begin{figure}[ht]
\centering
\includegraphics[width=\linewidth]{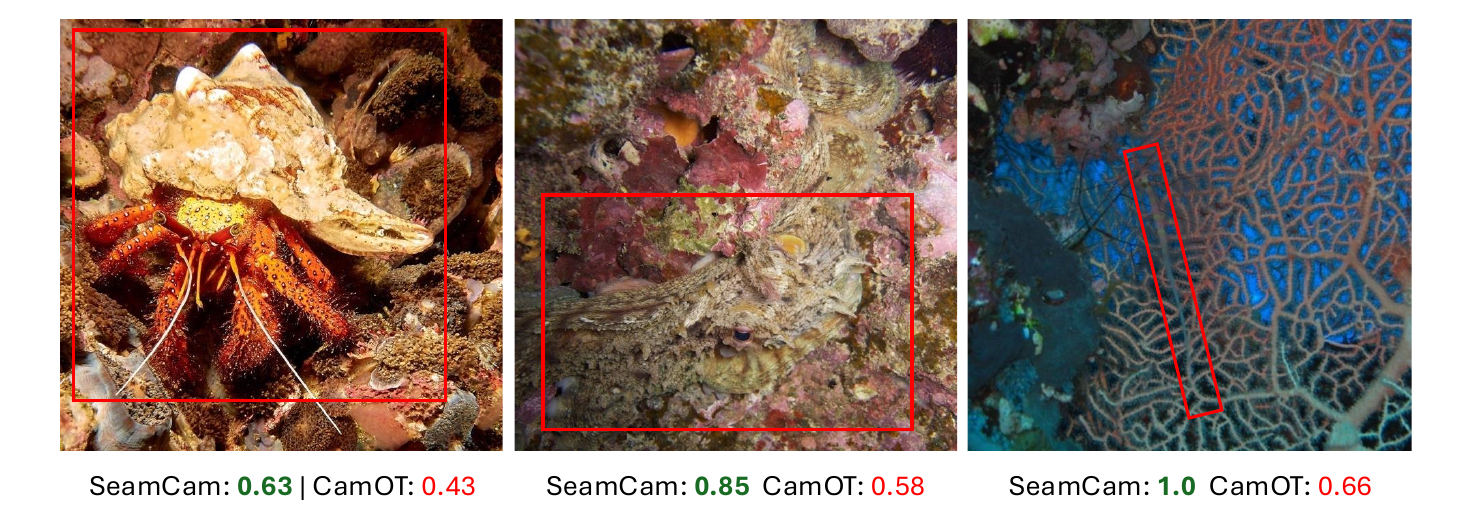}
\caption{\textbf{Disruptive coloration is correctly rewarded by \ourmethod.} Animals
relying on broken-outline strategies receive high \ourmethod\ scores
because the disruption fragments detection proposals onto pattern-matched background
regions, reducing the maximum recoverable localization signal. Similarity-based metrics
that compare foreground and background appearance statistics tend to under-score these
cases.}
\label{fig:appendix:disruptive}
\end{figure}

\paragraph{\textbf{Qualitative Evidence.}} Figure~\ref{fig:appendix:disruptive} shows
three examples of animals relying on disruptive coloration. \ourmethod\ assigns high
scores (0.63, 0.85, and 1.0 respectively) to these cases, consistent with their
perceptual difficulty. Similarity-based metrics would tend to under-score the
high-contrast cases due to the large internal feature variance relative to the
background.

\paragraph{\textbf{Aggregate Confirmation.}} The per-species accuracy comparison in
Figure~5 of the main paper provides aggregate confirmation: species that commonly rely
on disruptive patterning (e.g., \textit{Butterfly}, \textit{Spider}, \textit{Lizard})
are precisely those where \ourmethod's margin over CamOT is largest, with accuracy gaps
exceeding 0.20 in several cases.

\section{Biome Coverage and Cross-Texture Consistency}
\label{sec:appendix:biome}

Because \ourmethod\ is grounded in localization difficulty rather than in
foreground--background distance, background texture statistics do not enter the score
computation. Concretely, a cluttered high-entropy forest in which an animal blends into
the surrounding texture and a low-entropy desert in which boundary cues are weak both
produce low recoverability $D$ and therefore the same high \ourmethod\ score, despite
having radically different background statistics. The metric cannot be ``inconsistent''
across a variable it does not consume.

\paragraph{\textbf{Empirical Biome Distribution.}} To verify that the 2AFC evaluation
set covers a broad range of habitats, we ran open-vocabulary CLIP~\cite{monsefi2024detailclip} scene classification on
the full set and grouped the predictions into five biome buckets. The resulting
distribution is reported in Table~\ref{tab:appendix:biome}.

\begin{table}[ht]
\centering
\small
\setlength{\tabcolsep}{6pt}
\caption{\textbf{Biome distribution of the 2AFC evaluation set} based on CLIP
open-vocabulary scene classification. The set covers all major biome types.}
\label{tab:appendix:biome}
\begin{tabular}{lccccc}
\toprule
Biome & Forest & Desert & Aquatic & Grassland & Other \\
\midrule
Proportion (\%) & 24.9 & 11.6 & 37.3 & 23.0 & 3.2 \\
\bottomrule
\end{tabular}
\end{table}

\paragraph{\textbf{Per-Habitat Consistency.}} The 20 species in Figure~5 of the main
paper collectively span aquatic (\textit{Fish}, \textit{Pipefish}, \textit{Sea Horse},
\textit{Scorpion Fish}, \textit{Ghost Pipefish}), forest (\textit{Spider},
\textit{Mantis}, \textit{Katydid}, \textit{Chameleon}, \textit{Lizard}), grassland and
shrub (\textit{Grasshopper}, \textit{Caterpillar}, \textit{Butterfly}, \textit{Moth}),
and mixed/wetland (\textit{Frog}, \textit{Toad}, \textit{Owl}, \textit{Bird}) habitats.
\ourmethod\ outperforms CamOT in \emph{every} category, with no habitat-specific failure
mode. This pattern is consistent with the metric being driven by localization difficulty
rather than by texture statistics.

\section{Detector Behavior: Occlusion versus Camouflage}
\label{sec:appendix:detector_behavior}

A potential concern is whether IoU-based detectability scores conflate camouflage with
simple occlusion: if the open-vocabulary detector and SAM-2 systematically failed on
partially occluded subjects, the resulting score increase would not reflect genuine
camouflage strength. We address this concern both empirically and structurally.

\begin{figure}[ht]
\centering
\includegraphics[width=0.95\linewidth]{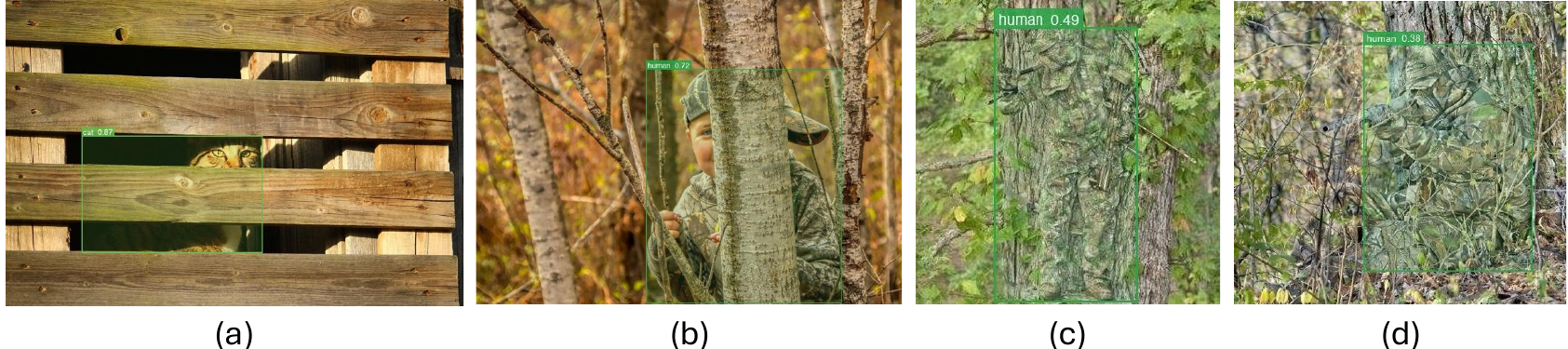}
\caption{\textbf{Detector confidence under occlusion versus camouflage.}
\emph{Left two images:} partially occluded but visually distinct animals are still
detected with high confidence. \emph{Right two images:} unoccluded but strongly
camouflaged animals receive weak detection confidence. The score reflects camouflage
effectiveness rather than occlusion, supporting that IoU-based detectability is not
dominated by occlusion-induced detector failure.}
\label{fig:appendix:occlusion}
\end{figure}

\paragraph{\textbf{Empirical Decoupling.}} Figure~\ref{fig:appendix:occlusion} illustrates
the decoupling. The first two examples show animals that are partially occluded but
visually salient: they are nevertheless detected with high confidence, producing low
\ourmethod\ scores. The second two examples show animals that are unoccluded and fully
visible, yet strongly camouflaged: they receive low detection confidence and weak
proposals, producing high \ourmethod\ scores. \ourmethod\ therefore tracks
\emph{detectability under category-conditioned search}, not visibility reduction by
occlusion.

\paragraph{\textbf{Structural Reasons.}} Three pipeline-level properties reduce the risk
of occlusion-induced confounding. First, Grounding DINO and SAM-2 are trained on object
distributions that include substantial partial occlusion, so occlusion alone does not
systematically defeat proposal generation. Second, the permissive confidence threshold
$\tau_\beta{=}0.10$ is specifically tuned to retain weak-but-valid cues from
partially-visible animals --- exactly the regime where camouflage and occlusion can be
confused. Third, the subset-max aggregation means that even partial detections from
non-occluded regions of the animal can contribute to $D$. Taken together with the
empirical 78.82\% agreement with human judgments over 2{,}290 pairs --- a
$\sim$25-point margin over CamOT --- these properties indicate that detector failure
under occlusion is not a dominant source of variance in the metric.

\section{Runtime and Computational Cost}
\label{sec:appendix:runtime}

We report a detailed runtime breakdown for \ourmethod\ and contextualize the cost against
existing camouflage metrics.

\paragraph{\textbf{End-to-End Runtime.}} On a single NVIDIA A100 GPU,
\ourmethod\ runs end-to-end at approximately \textbf{0.565\,s/image}. CamOT, which
operates on CPU only, runs at approximately \textbf{0.32\,s/image}. \ourmethod\ is
therefore less than $2\times$ slower than CamOT while improving human alignment by
$\sim$25 points. For evaluation metrics, perceptual alignment with ground truth is
typically the limiting factor, and the field has accepted slower but more accurate
evaluators in similar contexts (e.g., HPS-v2, LLM-as-judge, and GPT-4V-based scoring).

\paragraph{\textbf{Per-Stage Cost.}} The bulk of the runtime is spent in (i) the
open-vocabulary detection step and (ii) the segmentation step. Subset enumeration, often
flagged as a potential bottleneck because it scales as $2^K - 1$, is not in practice the
limiting factor: with $K$ capped at 7 (Section~\ref{sec:appendix:hyperparams}), at most
127 subset evaluations are required per image, each consisting of bitmask union and IoU
computation on precomputed masks. As shown in Figure~\ref{fig:seamcam_hyperparam}(a),
per-image runtime grows from 0.556\,s at $K{=}1$ to 0.565\,s at $K{=}12$ --- the subset
enumeration step contributes less than 2\% of total latency.

\paragraph{\textbf{Practicality.}} Both \ourmethod\ and CamOT are practical for offline
scoring at dataset scale. Given that the alignment gap is substantial and the cost gap is
modest, the trade-off favors \ourmethod\ for any setting where camouflage quality is the
primary axis of evaluation.

\section{Downstream Utility for Camouflaged Object Detection}
\label{sec:appendix:cod}

\ourmethod\ is designed as a camouflage \emph{evaluation} metric, and its primary
downstream application in this paper is preference-based fine-tuning of a generative
inpainting model (Section~\ref{exp:dpo_as_prefer}). A natural follow-up question is
whether the metric also provides value for camouflaged object detection (COD). This
section reports a direct experiment that supports this and discusses additional uses.

\paragraph{\textbf{Zero-Shot COD Evaluation on Generated Images.}} As an independent
external check on the camouflage strength of generated images, we evaluated outputs from
several camouflage-generation methods using a pretrained COD model
(DGNet~\cite{ji2023deep}) in zero-shot mode on \ourdataset. Under this protocol, lower DGNet IoU on
generated images indicates harder-to-detect camouflage from the perspective of an
external detector that has no exposure to either our metric or our generator. As
reported in Table~\ref{tab:appendix:cod}, \ourmethod-guided DPO produces images that
the independent DGNet model finds substantially harder to detect than those from prior
generators.

\begin{table}[ht]
\centering
\small
\setlength{\tabcolsep}{8pt}
\caption{\textbf{Zero-shot COD evaluation on \ourdataset.} A pretrained DGNet model
(zero-shot, never exposed to \ourmethod\ or our DPO model) is run on the generated
images. Lower IoU indicates stronger camouflage. \ourmethod-guided DPO yields the lowest
DGNet IoU, confirming that the improvements measured by \ourmethod\ transfer to an
independent COD model.}
\label{tab:appendix:cod}
\begin{tabular}{lc}
\toprule
Generator & DGNet IoU $\downarrow$ \\
\midrule
LAKE-RED~\cite{zhao2024lake}        & 0.7529 \\
TFill~\cite{zheng2022bridging}      & 0.6443 \\
\textbf{DPO (\ourmethod)}           & \textbf{0.5187} \\
\bottomrule
\end{tabular}
\end{table}

This zero-shot transfer is important because the evaluator (DGNet) is independent from
the scorer (\ourmethod) used during DPO training, ruling out the in-distribution reward
hacking pattern observed for DPO (CamOT) in Table~\ref{tab:dpo_ablation}. Combined with
the improved KID/FID/HPS-v2 numbers in the main paper, this provides strong evidence
that \ourmethod-guided DPO produces \emph{genuinely} harder camouflage rather than images
that merely score well under the training-time metric.

\paragraph{\textbf{Additional COD Use Cases.}} Beyond providing harder training data,
\ourmethod\ supports standard COD training and benchmarking recipes because $\zeta$
directly quantifies the same localization difficulty COD models must solve. This makes
the metric a natural drop-in for: (i) \emph{hard-example mining}, where training samples
are weighted by their \ourmethod\ score so that high-camouflage instances contribute more
to the loss; (ii) \emph{curriculum learning}, where samples are ordered from low to high
\ourmethod\ to stabilize early training; and (iii) \emph{difficulty-aware evaluation},
where COD benchmark performance is stratified by \ourmethod\ score to expose
failure modes that aggregate metrics obscure. We view these as concrete avenues for
applying \ourmethod\ within the broader COD ecosystem.
\end{document}